# Improving the Accuracy of Pre-trained Word Embeddings for Sentiment Analysis

Seyed Mahdi Rezaeinia [a,b], Ali Ghodsi [a], Rouhollah Rahmani [b]

[a] Department of Statistics and Actuarial Science, University of Waterloo, Waterloo, Canada
[b] Network Science and Technology Department, University of Tehran, Tehran, Iran

**Abstract**

Sentiment analysis is one of the well-known tasks and fast growing research areas in natural language processing (NLP) and text classifications. This technique has become an essential part of a wide range of applications including politics, business, advertising and marketing. There are various techniques for sentiment analysis, but recently word embeddings methods have been widely used in sentiment classification tasks. Word2Vec and GloVe are currently among the most accurate and usable word embedding methods which can convert words into meaningful vectors. However, these methods ignore sentiment information of texts and need a huge corpus of texts for training and generating exact vectors which are used as inputs of deep learning models. As a result, because of the small size of some corpuses, researcher often have to use pre-trained word embeddings which were trained on other large text corpus such as Google News with about 100 billion words. The increasing accuracy of pre-trained word embeddings has a great impact on sentiment analysis research. In this paper we propose a novel method, Improved Word Vectors (IWV), which increases the accuracy of pre-trained word embeddings in sentiment analysis. Our method is based on Part-of-Speech (POS) tagging techniques, lexicon-based approaches and Word2Vec/GloVe methods. We tested the accuracy of our method via different deep learning models and sentiment datasets. Our experiment results show that Improved Word Vectors (IWV) are very effective for sentiment analysis.

## 1. Introduction

Sentiment analysis is a practical technique that allows businesses, researchers, governments, politicians and organizations to know about people's sentiments, which play an important role in decision making processes. Word Embedding is one of the most useful deep learning methods used for constructing vector representations of words and documents. These methods have received a lot of attention in text and sentiment analysis because of their abilities to capture the syntactic and semantic relations among words. The two successful deep learning methods of word embeddings are Word2Vec [1,2] and Global Vectors (GloVe) [3]. Many researchers have used these two methods in their sentiment analysis research [4,5,6,7].

Although very effective, these methods have several limits and need to be improved. The Word2Vec and GloVe need very large corpuses for training and presenting an acceptable vector for each word [8,6]. For example, Google has used about 100 billion words for training Word2Vec algorithms and has re-released pre-trained word vectors with 300 dimensions. Because of the small size of some datasets, investigators have to use pre-trained word vectors such as Word2Vec and GloVe, which may not be the best fit for their data [9,10,11,12,13,14,15]. Another problem is that the word vector calculations of the two methods that are used to represent a document do not consider the context of the document [16]. For example, the word vector for "beetle" as a car is equal to its word vector as an animal. In addition, both models ignore the relationships between terms that do not literally co-occur [16]. Also, Cerisara et al. [17] have found that the standard Word2Vec word embedding techniques don't bring valuable information for dialogue act recognition in three different languages. Another important problem of these word embedding techniques is ignoring the sentiment information of the given text [6,7,8]. The side effect of this problem is that those words with opposite polarity are mapped into close vectors and it is a disaster for sentiment analysis [4].

In this research, we propose a novel method to improve the accuracy of pre-trained Word2Vec/Glove vectors in sentiment analysis tasks. The proposed method was tested by different sentiment datasets and various deep learning models from published papers. The results show that the method increases the accuracy of pre-trained word embeddings vectors for sentiment analysis. The organization of this paper is as follows: Section 2 describes the related works and literature review for this research. Section 3 presents our proposed method and algorithm, and additionally describes the proposed deep learning model for testing the method. Section 4 reports our experiments, showing results along with evaluations and discussions. Section 5 is the conclusion and future works.

## 2. Related Works

### 2.1. Lexicon-based method

Sentiment classification techniques are mainly divided into lexicon-based methods and machine learning methods such as Deep Learning [18,19]. The lexicon-based sentiment analysis approach is typically based on lists of words and phrases with positive and negative connotations [20,21,22]. This approach needs a dictionary of negative and positive sentiment values assigned to words. These methods are simple, scalable, and computationally efficient. As a result, they are mostly used to solve general sentiment analysis problems [18]. However, lexicon-based methods depend on human effort in human-labeled documents [19] and sometimes suffer from low coverage [8]. Also, they depend on finding the sentiment lexicon which is applied to analysis the text [18].

Because of the accuracy improvement of text classification, the lexicon-based approaches have combined with machine learning methods recently. Several authors found that the machine learning methods were more accurate than the lexicon methods [19,23]. Mudinas et al. [24] increased the accuracy of sentiment analysis by combining lexicon-based and Support Vector Machine (SVM) methods. Zhang et al. [25] successfully combined lexicon-based approach and binary classifier for sentiment classification of Twitter data. Basari et al. [26] have combined the Particle Swarm Optimization (PSO) technique and SVM method for sentiment analysis of movie reviews. In all of these cases, the machine learning techniques improved the accuracy of text classifications.

## 2.2. Deep learning method

Recently, deep learning methods have played an increasing role in natural language processing. Most of the deep learning tasks in NLP has been oriented towards methods which using word vector representations [6]. Continuous vector representations of words algorithms such as Word2Vec and GloVe are deep learning techniques which can convert words into meaningful vectors. The vector representations of words are very useful in text classification, clustering and information retrieval. Word embeddings techniques have some advantages compare to bag-of-words representation. For instance, words close in meaning are near together in the word embedding space. Also, word embeddings have lower dimensionality than the bag-of-words [2].

The accuracy of the Word2vec and Glove depends on text corpus size. Meaning, the accuracy increases with the growth of text corpus. Tang et al. [4] proposed learning continuous word representations for sentiment analysis on Twitter which is a large social networks dataset. Severyn and Moschitti [27] used Word2Vec method to learn the word embeddings on 50M tweets and applied generated pre-trained vectors as inputs of a deep learning model. Recently, Lauren et al. [28] have proposed a discriminant document embeddings method which has used skip-gram for generating word embeddings of clinical texts. Fu et al. [5] applied Word2Vec for word embeddings of English Wikipedia dataset and Chinese Wikipedia dataset. The word embeddings used as inputs of recursive autoencoder for sentiment analysis approach. Ren et al. [7] proposed a new word representation method for Twitter sentiment classification. They used Word2Vec to generate word embeddings of some datasets in their method. Qin et al. [29] trained Word2Vec algorithm by English Wikipedia corpus which has 408 million words. They used those pre-trained vectors as inputs of convolutional neural networks for data-driven tasks.

Nevertheless, as already mentioned, these word embedding algorithms need a huge corpus of texts for training [8] and most of them ignore the sentiment information of text [4,6]. Because of the limitations and restrictions in some corpuses, researchers prefer to use pre-trained word embeddings vectors as inputs of machine learning models. Kim [9] has used pre-trained Word2Vec vectors as inputs to convolutional neural networks and has increased the accuracy of sentiment classification. Also, Camacho-Collados et al. [11] used pre-trained Word2Vec vectors for the representation of concepts. Zhang and Wallace [10] have applied pre-trained GloVe and Word2Vec vectors in their deep learning model and have improved the accuracy of sentence and sentiment classification. Caliskan et al. [12] used pre-trained GloVe word embeddings vectors for increasing the accuracy of their proposed method. Wang et al. [13] applied pre-trained GloVe vectors as inputs of attention-based LSTMs model for aspect-level sentiment analysis. Liu et al. [15] have used pre-trained Word2Vec as a word embedding representation for recommending Idioms in essay writing.

As a result, increasing the accuracy of pre-trained word embedding is very important and plays a vital role in sentiment classification methods. Zhang and Wallace [10] combined pre-trained Word2Vec and GloVe vectors in their deep learning model, but the accuracies were decreased. Some results are shown in table 1.

Table1: The results of the combination of pre-trained Word2Vec and Glove on three datasets [10]

| Dataset | Reduced accuracy rate (%) |
|---|---|
| Movie reviews (MR) | **-0.22 %** |
| Stanford sentiment treebank (SST) | **-1.1%** |
| TREC | **-0.17%** |

According to table1, combination of pre-trained Word2Vec and Glove decreased the accuracy of text and sentiment classification on some datasets. Also, Kamkarhaghighi and Makrehchi [16] have proposed an algorithm for increasing the accuracy of pre-trained Word2Vec and Glove. Their algorithm was tested on two datasets and the accuracy of Word2Vec was decreased on one dataset by the proposed algorithm. In the following section we present in detail our proposed model, its algorithm and the proposed deep learning model for checking our method.

## 3. Proposed method

In our proposed method, Improved Word Vector (IWV) we have increased the accuracy of word embedding vectors based on the combination of natural language processing techniques, lexicon-based approaches and Word2Vec/GloVe methods which have high accuracies. The main architecture of the proposed method has been shown in figure 1.

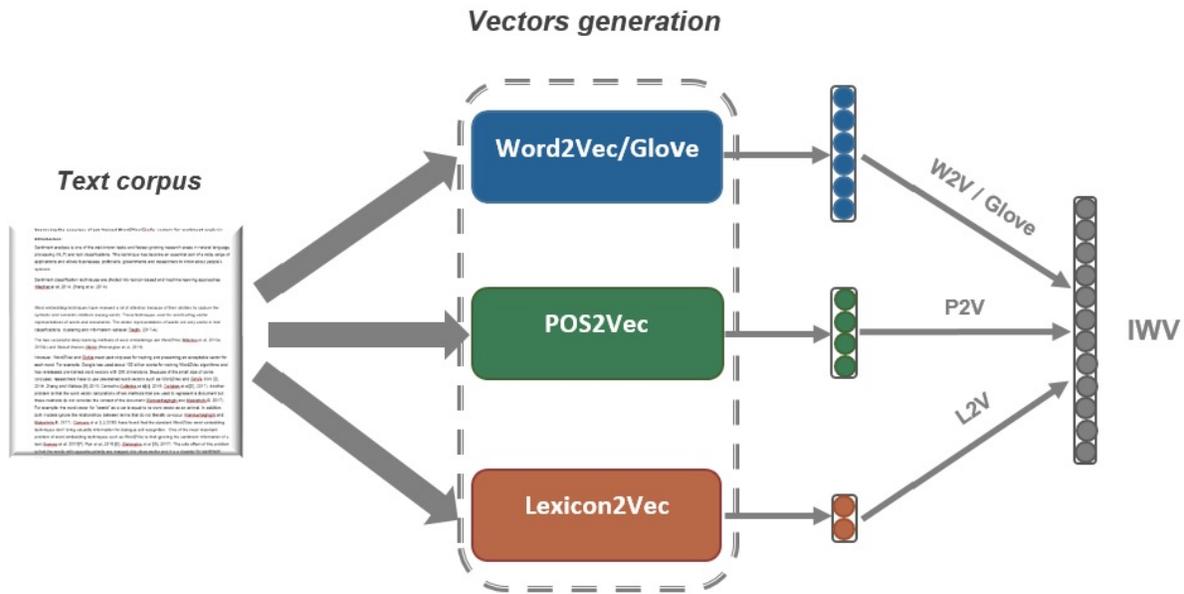

Figure1: The main architecture of the proposed method (Improved word vector)

### 3.1. Word2Vec and GloVe

Word2Vec and GloVe are two successful word embedding algorithms which have high accuracy. Word2Vec is based on continuous Bag-of-Words (CBOW) and Skip-gram architectures which can provide high quality word embedding vectors. CBOW predicts a word given its context and Skip-gram can predict the context given a word. The generated vectors of words which appear in common contexts in the corpus are located close to each other in the vector space. GloVe word embedding is a global log-bilinear regression model and is based on co-occurrence and factorization of matrix in order to get vectors. In this study, we used 300-dimension word2vec embeddings trained on Google News and 300-dimension GloVe word embeddings.

## 3.2. POS2Vec (P2V)

Part-of-speech (POS) tagging is an important and fundamental step in Natural Language Processing which is the process of assigning to each word of a text the proper POS tag. The Part-of-speech gives large amount of information about a word and its neighbors, syntactic categories of words (nouns, verbs, adjectives, adverbs, etc.) and similarities and dissimilarities between them. We converted each generated POS tag to a constant vector and concatenated with Word2Vec/GloVe vectors. We assigned 50-dimensional vectors to each word. As a result, Word2Vec/Glove vectors will have syntactic information of words.

## 3.3. Lexicon2Vec (L2V)

The sentiment and emotion lexicons are lists of phrases and words which have polarity scores and can be used to analyze texts. Each lexicon contains of words and their values which are the sentiment scores for that words. There are various sentiment and emotion lexicons that can be used, but choosing the proper lexicon is very important. We selected six lexicons as our resources and assigned 6-dimensional vectors to each word.

- National Research Council Canada (NRC) Emoticon Affirmative Context Lexicon and NRC Emoticon Negated Context Lexicon [30,31,32].
- NRC Hashtag Affirmative Context Sentiment Lexicon and NRC Hashtag Negated Context Sentiment Lexicon [30,31,32].
- NRC Emoticon Lexicon [30,31,32].
- NRC Hashtag Sentiment Lexicon [30,31,32].
- SemEval-2015 English Twitter Sentiment Lexicon [33,31].
- Amazon Laptop Sentiment Lexicon [34].

Finding the proper combination of lexicons is not easy and plays a vital role in sentiment analysis. We tested various lexicons and selected six of them as one of the best combination.

Algorithm 1 shows the process of our proposed method. It gets a sentence and returns improved word vectors of the sentence. In the first step, assign a constant vector to each POS tag. In the second step, each word vector of the input sentence is extracted from Word2Vec or GloVe datasets and if a word doesn't exist in the datasets its vector will generate randomly. In the third step, POS tag of each word is determined and assign a constant vector to each one. In the next step, sentiment scores of each word are extracted from all lexicons and will normalize them. If a word doesn't exist in any lexicons, its score will be zero. The generated vectors from each step will be concatenated with other vectors from previous steps.

**Algorithm 1 : Improved Word Vector (IWV) generation**

**Inputs:**

S = {W$_1$, W$_2$,……., W$_n$} , *Input sentence S contains n words*
PT = {T$_1$, T$_2$,……, T$_m$}, *All POS tags*
W2VD: *Google Word2Vec Dataset*
GloVeD: *Glove Dataset*
h: Number of lexicons
LEX$_1$, LEX$_2$,……., LEX$_h$     *All lexicons*

**Output:**

IMV: *Improved word vectors of sentence S*

1. **for** j=1 **to** m **do**
2.   VT$_j$ ⟵ *GenerateVector* ( T$_j$ )
3.   T$_j$ ⟵ < T$_j$ , VT$_j$ >
4. **end for**

5. **for each** W$_i$ **in** S **do**
6.     **If** W$_i$ **exist in** W2VD **then** *extract* VecW$_i$
7.         MV$_i$ ⟵ VecW$_i$
8.     **elseif** W$_i$ **exist in** GloVeD **then** *extract* VecW$_i$
9.         MV$_i$ ⟵ VecW$_i$
10.    **else**
11.        MV$_i$ ⟵ *RandomGenVec* (VecW$_i$)
12.    **endif**

13.    POS ⟵ *ExtractPOS* ( W$_i$ )
14.    **for** k=1 **to** m **do**
15.        **If** POS=T$_k$ **then**  **ADD** VT$_k$ **into** MV$_i$
16.        **end if**
17.    **end for**

18.    **for** k=1 **to** h **do**
19.        **If** W$_i$ **in** LEX$_k$ **then**
20.            S$_{ik}$ ⟵ *FindVector* ( W$_i$ )
21.            S$_{ik}$ ⟵ *Normalize* ( S$_{ik}$ ) *Between -0.995 and +0.995*
22.        **else**
23.            S$_{ik}$ ⟵ 0
24.        **end if**
25.        **ADD** S$_{ik}$ **into** MV$_i$
26.    **end for**
27.    **ADD** MV$_i$ **into** IMV
28.  **Return** IMV
29. **end for**

We proposed a deep learning model (we called it Model 1) for evaluating our generated vectors on well-known datasets. The model consists of three convolutional neural networks (CNN), a pooling, and a fully connected layer and the inputs of the model are the improved word vectors (IWV). The model is shown in figure 2.

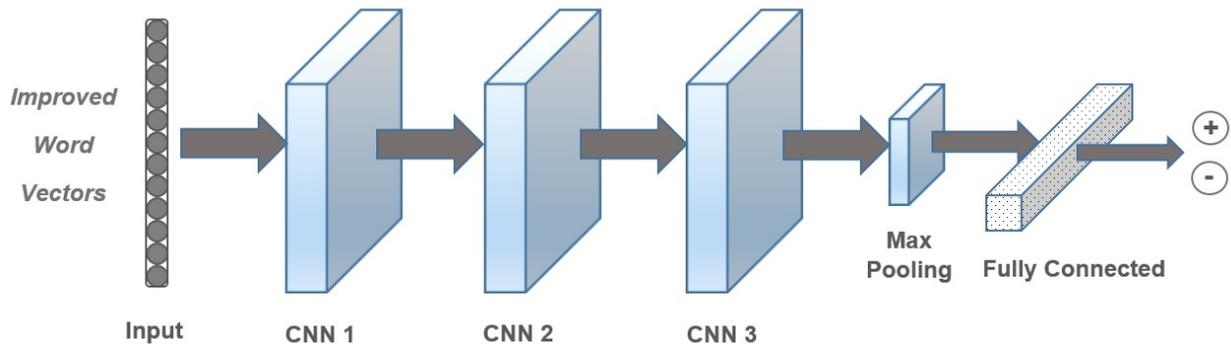

Figure 2: The proposed model (Model1) for evaluating our proposed approach

In addition, for more confidence about the accuracy of our method, we checked the proposed method by other three deep learning models from published papers. The second deep learning model (Model2) was introduced by Kim [9]. The third model (Model3) was applied by Ouyang et al. [35] and the fourth one (Model4) was proposed by Deriu et al. [36].

## 4. Experiments

In this section, we describe the datasets and experimental evaluations to show the effectiveness of our proposed method.

### 4.1. Datasets

Datasets that were used in our research are as follows:

**MR:** Movie reviews dataset with equal positive and negative sentences and each review contains a sentence [37].

**CR:** Customer reviews of products classified into positive and negative reviews [22].

**SST:** The Stanford sentiment treebank introduced by Socher et al. [38] contains train/development/test texts. We used only train and test sentences and binary labels (positive, negative) for our model.

### 4.2. Results

We have tested our approach on four different deep learning models and various sentiment datasets which have different features. Our implementations were GPU-based and have trained our models on four GeForce GTX Titan X GPUs. We have used Tensorflow for implementing and training all the deep learning models in our research. All reports are based on the average of accuracies calculated over multiple runs of 10-fold cross-validation(CV), however SST has predefined train and test sets. In 10-fold CV, the dataset is randomly partitioned into 10 folds. Among them, nine folds are used for training and the remaining one fold is used for testing. We compared the results of the 356-dimension IWV with 300-dimension Word2Vec and 300-dimension GloVe.

As already mentioned before, six sentiment lexicons were used to extract and generate the lexicon vectors. We only used unigram scores for our research. The distribution of the lexicons used in our research is listed in table 2.

| Lexicon | Positive | Negative | Neutral | Total | Scores Ranges |
|---|---|---|---|---|---|
| NRC Emoticon Affirmative Context Lexicon and NRC Emoticon Negated Context Lexicon | 28025 | 27121 | 0 | 55146 | -5.844 to +4.495 |
| NRC Hashtag Affirmative Context Sentiment Lexicon and NRC Hashtag Negated Context Sentiment Lexicon | 19502 | 24447 | 0 | 43949 | -10.025 to +10.661 |
| NRC Emoticon Lexicon | 38312 | 24156 | 0 | 62468 | -4.999 to +5.0 |
| NRC Hashtag Sentiment Lexicon | 32048 | 22081 | 0 | 54129 | -6.925 to +7.526 |
| SemEval-2015 English Twitter Sentiment Lexicon | 776 | 726 | 13 | 1515 | -0.984 to +0.984 |
| Amazon Laptop Sentiment Lexicon | 14651 | 11926 | 0 | 26577 | -5.27 to +3.702 |

Table 2: Statistics of the lexicons which were used in the research

In our proposed deep learning model the filter size of CNN1, CNN2 and CNN3, respectively are 3,4 and 5 with 100 feature maps each. The flatten layer has 95 nodes and the activation function of the layers is Rectified Linear Unit (ReLU). The results are shown in table 3.

Table 3: The accuracy (%) comparisons between our method (IWV) and other methods based on the deep learning model 1

| Method | Dim | MR | CR | SST |
|---|---|---|---|---|
| Word2Vec | 300 | 79.3 | 81.8 | 82.0 |
| GloVe | 300 | 79.2 | 81.3 | 81.0 |
| IWV | 356 | **79.8** | **82.5** | **83.7** |

As shown in table3, the accuracy of our combined vector is higher than the existing pre-trained vectors on three sentiment datasets. In other word, the proposed method increased the accuracy of sentiment analysis in our proposed deep learning model. The results show that the accuracy of SST, CR and MR were increased by 1.7% ,0.7% and 0.5% respectively. Levy et al. [39] showed that Word2Vec performs better than GloVe in various tasks. Also, we found that the Word2Vec is generally more accurate than the GloVe, so all words are searched firstly on the Word2Vec dataset and then are searched on GloVe by the proposed algorithm.

In order to test the IWV more fully, we checked our approach by other three deep learning models on MR and CR datasets which are balanced and unbalanced. The results have shown in table 4.

Table 4: The accuracy (%) comparisons between our method and other methods based on models 2,3 and 4. Each cell reports the average accuracies calculated over multiple runs of 10-fold CV.

| Model | Dataset | Word2Vec(300) | GloVe(300) | IWV(356) |
|---|---|---|---|---|
| Model 2 | MR | 79.4 | 78.7 | **79.8** |
| Kim [9] | CR | 82.8 | 82.9 | **83.2** |
| Model 3 | MR | 78.0 | 77.3 | **78.2** |
| Ouyang et al. [35] | CR | 80.0 | 79.5 | **81.1** |
| Model 4 | MR | 79.4 | 78.4 | **79.6** |
| Deriu et al. [36] | CR | 81.6 | 80.1 | **81.8** |

According to table 4, our proposed method is more accurate than other methods in models 2, 3 and 4. As a result, the IWV was compared to other methods six times and all accuracies were improved.

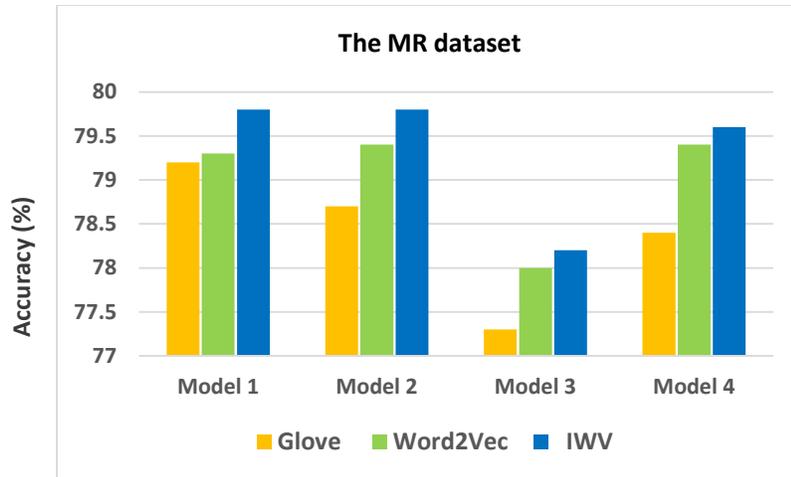

Figure 3: The accuracy (%) comparisons of three methods on four deep learning models for MR dataset

Figure 3 shows the accuracy of Glove, Word2Vec and IWV vectors on four deep learning models for MR which is a balanced dataset. It contains a total of 10662 reviews are divided into equal positive and negative reviews. As can be seen, IWV method has the highest accuracy and the Glove has the lowest accuracy among three methods. More generally, the IWV method has increased the accuracy of sentiment analysis in MR dataset between 0.2% and 0.5%.

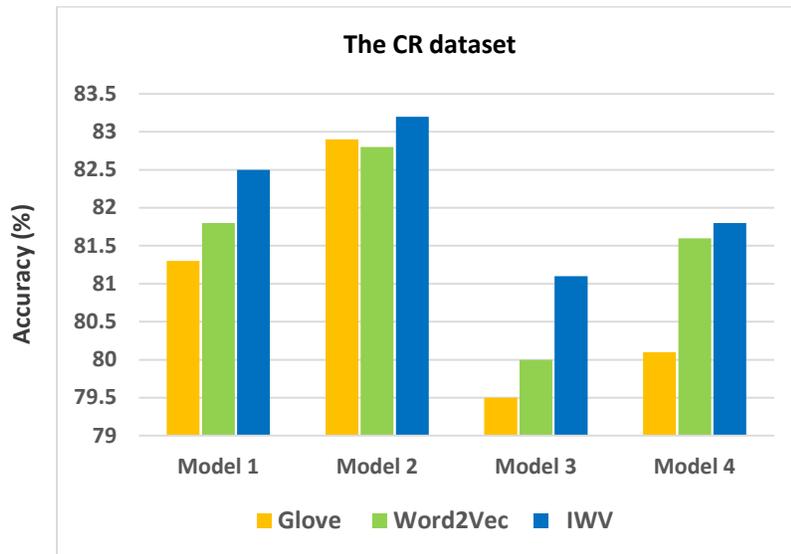

Figure 4: The accuracy (%) comparisons of three methods on four deep learning models for CR dataset

Figure 4 indicates that the IWV method generally has performed better than other per-trained word embeddings for sentiment analysis of CR dataset. The CR is an unbalanced dataset which contains 2397 positive and 1406 negative reviews. As can be seen, pre-trained Word2vec embedding is almost more accurate than pre-trained Glove embedding, however it is reverse in the model 2. The IWV provides absolute accuracy improvements of 0.7%, 0.4%, 1.1% and 0.2% for model 1, model 2, model 3 and model 4, respectively.

## 5. Conclusion

In this paper, we proposed a new method to improve the accuracy of well-known pre-trained word embeddings for sentiment analysis. Our method has improved the accuracy of pre-trained word embeddings based on the combination of three approaches such as lexicon-based approach, POS tagging approach and Word2Vec/GloVe approach. In order to ensure about the accuracy of our proposed method, we have tested it nine times with different deep learning models and sentiment datasets. The experimental results indicated that our method has increased the accuracy of sentiment classification tasks in all models and datasets. Briefly, the main advantages of the proposed method are:

- Because of the accuracy of pre-trained Word2Vec/Glove, adding any vector to them decreased the accuracy according to previous researches, but our proposed method has increased the accuracy of pre-trained vectors in sentiment analysis for the first time.

- One of the best combination of lexicons was introduced in our research. This combination increased the accuracy of all tested deep learning models and datasets.

- Any improvements in pre-trained word embeddings/POS tagging/Lexicons in the future, will increase the accuracy of our method.

As a result, our proposed method can be the basis for all sentiment analysis techniques which are used deep learning approaches.